\documentclass[runningheads]{llncs}
\usepackage[T1]{fontenc}
\usepackage{amsmath}
\usepackage{graphicx}
\usepackage{amsmath,amssymb}
\usepackage{algorithm}
\usepackage{algpseudocode}
\usepackage{listings}
\usepackage{enumitem}
\usepackage{multirow}
\usepackage{cite}

\begin{document}
%

\title{SEval-\includegraphics[height=1em]{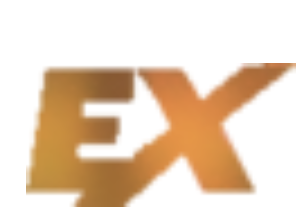}: A Statement-Level Framework for Explainable Summarization Evaluation}
\titlerunning{SEval-Ex}
%
\author{Tanguy Herserant\inst{1}\orcidID{0009-0002-0468-342X} \and \\
Vincent Guigue\inst{1}\orcidID{0000-0002-1450-5566}}

\authorrunning{T. Herserant and V. Guigue}
%
\institute{AgroParisTech - MIA, 22 place de l’Agronomie, 91120 Palaiseau, France \\
\email{\{tanguy.herserant, vincent.guigue\}@agroparistech.fr}}

\maketitle              
%

\begin{abstract}
Evaluating text summarization quality remains a critical challenge in Natural Language Processing. Current approaches face a trade-off between performance and interpretability.
We present SEval-Ex, a framework that bridges this gap by decomposing summarization evaluation into atomic statements, enabling both high performance and explainability.
SEval-Ex employs a two-stage pipeline: first extracting atomic statements from text source and summary using LLM, then a matching between generated statements. Unlike existing approaches that provide only summary-level scores, our method generates detailed evidence for its decisions through statement-level alignments. Experiments on the SummEval benchmark demonstrate that SEval-Ex achieves state-of-the-art performance with 0.580 correlation on consistency with human consistency judgments, surpassing GPT-4 based evaluators (0.521) while maintaining interpretability. Finally, our framework shows robustness against hallucination.

\end{abstract}

\section{Introduction}

The evaluation of text generation has become a critical challenge in Natural Language Processing (NLP), particularly as Large Language Models (LLMs) revolutionize our ability to generate human-like text~\cite{brown2020language, touvron2023llama, jiang2023mistral}. While these advances have enabled unprecedented fluency in text generation, they have also highlighted a fundamental challenge: how can we reliably assess the factual accuracy of generated content while maintaining interpretability in our evaluation methods?

Traditional approaches to information extraction, particularly Named Entity Recognition (NER), have shown significant limitations in specialized contexts, often failing to achieve satisfactory performance across diverse domains. While NER excels at identifying explicit facts, its rigid structure struggles with the nuanced ways information can be expressed in natural language. Meanwhile, semantic metrics operating in latent space, such as BERTScore~\cite{zhang_bertscore_2020}, offer intriguing possibilities but face a crucial ambiguity: do they truly measure factual accuracy, or are they primarily capturing linguistic fluency?

Leveraging the advanced capabilities of LLMs presents an innovative avenue to address these challenges. LLMs exhibit a high level of competence in text reformulation and comprehension~\cite{shu2024rewritelm}. We propose utilizing LLMs for information extraction through a textual reformulation approach, bypassing the complexities and potential errors associated with mapping text to structured data formats. This methodology capitalizes on the inherent language understanding of LLMs to identify and align fundamental knowledge units, that we call atomic statement, within the text. Our approach is inspired by recent methodologies such as RAGAS~\cite{es2023ragas}, yet is specifically tailored to tackle the technical challenges associated with evaluating longer texts that necessitate detailed analysis.

We present SEval-Ex, a novel framework that decomposes summarization evaluation into atomic statements, enabling both high performance and interpretable assessment of the factuality of information (consistency). Our approach makes three key technical contributions:
\begin{itemize}
    \item A \textbf{Statement Extraction methodology} that leverages LLMs to decompose both source and summary texts into atomic statements
    \item  \textbf{Verdict Reasoning Pipeline:} We design a pipeline that provides interpretable evaluation decisions by matching and classifying atomic statements into True Positives (TP), False Positives (FP), and False Negatives (FN)
    \item A \textbf{comprehensive evaluation protocol} that demonstrates good robustness across various forms of hallucination (entity, event, and detail-level) validated through extensive experimentation
\end{itemize}

To validate the effectiveness of our approach, we conducted rigorous testing using the SummEval benchmark~\cite{fabbri2021summeval}. The results demonstrate that SEval-Ex achieves a state-of-the-art Spearman correlation coefficient of $0.580$ with human consistency judgments, outperforming GPT-4-based evaluators ($0.521$). Importantly, our framework maintains complete interpretability, a critical feature for applications where understanding the rationale behind evaluation decisions is essential. SEval-Ex demonstrates particular robustness on various types of factual inconsistencies, underscoring its value for applications that demand high reliability and transparency in content evaluation.

\begin{figure*}[ht!]
    \centering
    \includegraphics[width=\textwidth]{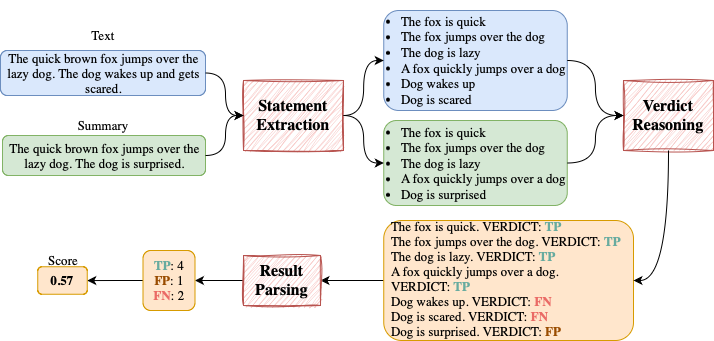}
    \caption{SEval-Ex evaluation pipeline. First, an LLM extract statements during the (1) Statement Extraction phase, then during (2) Verdict Reasoning phase, an LLM labels the statements. Finally, a (3) parser extract the confusion matrix that made the score.}
    \vspace{-0.5cm}
    \label{fig:model_architecture}
\end{figure*}

\section{Related Work}

The evaluation of summarization systems has evolved significantly, reflecting a shift from basic lexical matching techniques to advanced neural methodologies. In this section, we organize our discussion around five key paradigms: \textit{traditional lexical overlap metrics}, \textit{embedding-based approaches}, \textit{task-specific evaluation frameworks}, \textit{NLI-based methods}, and \textit{LLM-based evaluators}.
\subsubsection*{Traditional Lexical Overlap Metrics}
Early evaluation metrics for summarization, such as ROUGE~\cite{lin_rouge_2004} and BLEU~\cite{papineni2002bleu}, primarily relied on surface-level text matching by measuring \textit{n}-gram overlap between generated summaries and reference texts. While these metrics are computationally efficient and offer structural interpretability through visible lexical overlap, they exhibit significant limitations. Notably, they inadequately handle paraphrasing and synonymy, fail to capture deeper semantic equivalence, and overly depend on exact word matching.
\subsubsection*{Embedding-based Approaches}
The introduction of contextual embeddings, such as those from BERT~\cite{devlin2018bert}, facilitated the development of more sophisticated evaluation metrics that aim to capture semantic similarity beyond surface lexical overlap. Metrics like BERTScore~\cite{zhang_bertscore_2020} compute similarity scores based on token embeddings, while MoverScore~\cite{zhao2019moverscore} employs Earth Mover's Distance to compare distributions of embeddings between the candidate and reference texts. Despite these advancements, BERTScore has revealed significant limitations~\cite{hanna2021fine}. Specifically, embedding-based approaches may provide a coarse-grained semantic representation that overlooks nuanced differences, making them less sensitive to minor errors, especially when the candidate text is lexically or stylistically similar to the reference. This insensitivity can result in misleadingly high similarity scores despite critical differences in meaning.
\subsubsection*{Task-specific Evaluation Frameworks}
To address the unique challenges of summarization evaluation, specialized frameworks have been developed that focus on specific aspects such as factual consistency and overall quality. For instance, FactCC~\cite{kryscinski2019evaluating} employs natural language inference (NLI) models to detect factual inconsistencies between the summary and the source text. Similarly, QAGS~\cite{wang2020asking} and QuestEval~\cite{scialom2021questeval} utilize question answering techniques to verify factual accuracy by generating and answering questions derived from the summaries and source documents. 
%
Multi-aspect evaluation frameworks like UniEval~\cite{zhong_towards_2022} and SUPERT~\cite{gao2020supert} provide comprehensive assessments by evaluating dimensions such as fluency, coherence, and informativeness. However, they often face high computational costs, domain specificity limiting generalizability, and challenges in interpretability, making them less practical for widespread use.
\subsubsection*{NLI-based Approaches}
NLI-based approaches apply entailment models to assess the factual consistency between summaries and source documents. Methods such as SummaC~\cite{laban2022summac} evaluate whether the content of a summary can be inferred from the source text by computing entailment probabilities across sentence pairs. 
These approaches show promise in detecting factual inaccuracies but face significant challenges, particularly in scalability, due to the combinatorial complexity of computing entailment relations, making the process computationally expensive for longer texts.
Recent efforts by Liu et al.~\cite{liu2021natural} introduce contextual reasoning to mitigate some of these limitations, yet significant challenges persist in effectively adapting these models across different domains and handling the complexities inherent in long-form text understanding.
\subsubsection*{LLM-based Evaluators}
Leveraging the capabilities of LLMs, recent evaluation methods have begun to utilize models like GPT-3~\cite{brown2020language} and GPT-4 for assessing summarization quality. Techniques such as GPTScore~\cite{fu_gptscore_2023} and G-Eval~\cite{liu_g-eval_2023} employ LLMs to generate evaluation scores or comparative judgments between summaries. 
These methods leverage LLMs' extensive pretraining and nuanced language understanding but face challenges like their black-box nature, limiting transparency and interpretability, as well as high computational costs, making large-scale or real-time evaluations difficult.
Moreover, reliance on proprietary models raises concerns about reproducibility and accessibility~\cite{bommasani2021opportunities}.

\section{Methodology}

We present SEval-Ex, a framework that evaluates summarization through atomic statements—self-contained units of information—for a fully interpretable assessment. By prioritizing transparency at every stage, SEval-Ex allows users to trace evaluation decisions from statement extraction to final scoring. Unlike black-box approaches that provide only summary-level scores, our method delivers granular insights through three transparent steps: (1) Statement Extraction, which interprets source texts and summaries into atomic statements; (2) Verdict Reasoning, which aligns and classifies these statements (True Positives, False Positives, False Negatives); and (3) Results Parsing, which computes metrics based on these alignments.
To enhance efficiency and maintain context, statement verification is directly integrated into LLM prompts, avoiding exhaustive pairwise comparisons. This design ensures both scalability and interpretability. Figure~\ref{fig:model_architecture} illustrates the pipeline and highlights its core functionalities with an example.

\subsection{Problem Formalization}

Given a source document $D$ and its generated summary $S$, our goal is to assess the summary's quality through statement-level analysis. Let $\mathcal{S}$ be the space of all possible atomic statements and $\mathcal{Y} = \{\text{TP}, \text{FP}, \text{FN}\}$ be the verification outcome space. Let's formalize the key components:

\paragraph{Statement Extraction Function:} $E(\cdot)$ decomposes text into atomic statements:
\begin{align}
    E(D) &= \{d_1, d_2, \ldots, d_n\} \text{ where } d_i \in \mathcal{S} \text{ is an atomic statement from } D \\
    E(S) &= \{s_1, s_2, \ldots, s_m\} \text{ where } s_j \in \mathcal{S} \text{ is an atomic statement from } S
\end{align}

\paragraph{Statement Verification:} $V(\cdot,\cdot)$ assesses relationships between statements:
\begin{equation}
    V: E(D) \times E(S) \rightarrow \mathcal{Y}
\end{equation}

\paragraph{Score Function:} $F(\cdot,\cdot)$ computes the final evaluation metrics:
\begin{equation}
    F(D,S) = \text{F}_1(V(E(D), E(S))) \rightarrow \mathbb{R}
\end{equation}

\subsubsection*{Statement Extraction}
The statement extraction process uses quantized LLMs to decompose text into atomic statement.
We optimize prompts through extensive experimentation, the statement extraction prompt is designed to decompose text into atomic statements, which are self-contained facts.

\subsubsection*{Verdict Reasoning}

The verdict reasoning function $V$ evaluates the relationship between extracted statements as follows: 
%
\begin{itemize}
    \item True Positive (TP): $s_j = TP$ if $d_i \equiv s_j$
    \item False Positive (FP): $s_j = FP$ if $\nexists d_i: d_i \equiv s_j$
    \item False Negative (FN): $d_i = FN$ if $\nexists s_j: d_i \equiv s_j$
\end{itemize}

where $\equiv$ denotes semantic equivalence between statements. We define two statements as semantically equivalent ($\equiv$) when they convey the same fact despite linguistic differences, evaluated by a LLM.\\

The objective is to measures the presence of summary information in the text. A summary is considered correct if its main statements are supported by corresponding statements in the source text. To evaluate this, we compute precision, recall and F1 score. 

Obviously, we seek to verify the factual accuracy of the information, not to weigh its importance. Our goal is therefore only to assess the consistency of the summary, not its relevance or coherence.




\subsection{Pipeline Improvements for Long Texts}
\label{sec:var}

The statement extraction and matching process faces two fundamental challenges: (1) information loss during statement decomposition, where contextual relationships between sentences may be lost when extracting atomic statement, and (2) semantic drift during matching, where extracted statements may lose their original meaning when compared in isolation. These challenges are particularly acute for longer texts, where maintaining coherence across statement boundaries becomes increasingly difficult.

To address these challenges, we explore modifications to our base pipeline (Figure \ref{fig:model_architecture}) that aim to preserve contextual information while maintaining computational efficiency. Our investigation focuses on two key areas: optimizing the granularity of text processing during extraction, and simplifying the matching process to reduce semantic drift.

\begin{enumerate}
    \item \textbf{Statement Extraction}:
    \begin{itemize}
        \item \textbf{Base}: Processes the entire text as a single unit, using one LLM call. While efficient, this approach can lose local context in longer documents, particularly when statements reference information across distant sentences.
        \item \textbf{3-Chunk}: Segments the text into three-sentence chunks, enabling more focused local context preservation through multiple LLM calls. This granular approach better captures inter-sentence relationships while remaining computationally feasible.
    \end{itemize}
    
    \item \textbf{Verdict Reasoning}:
    \begin{itemize}
        \item \textbf{Base}: Performs exhaustive comparison between all extracted statements from text and summary. While comprehensive, this approach can introduce semantic drift as statements are matched without their surrounding context.
        \item \textbf{StSum\_Text}: Directly matches summary statements to source text sentences, reducing information loss by preserving the original textual context during verification.
        It is also a simple way to address the first issue regarding the longest text.
    \end{itemize}
    
\end{enumerate}

After extensive comparative experiments, we selected \texttt{Qwen2.5:72B} model (4bits quantization).
The complete implementation, including prompt templates and optimization code, will be made publicly available on Github\footnote{https://github.com/TanguyHsrt/seval-ex}.

\section{Experiments}

We evaluate our approach on the widely-used benchmarks of summarization evaluation:
\textbf{SummEval} \cite{fabbri2021summeval}: a comprehensive benchmark containing 1,600 summaries generated by 16 different summarization systems on CNN/DailyMail articles. Each summary is annotated with human ratings across four dimensions: coherence (the collective quality of all sentences), consistency (the factual alignment between the summary and the summarized source), fluency (the quality of individual sentences) and relevance (selection of important content from the source). Since our work focuses on atomic statements, the consistency 
metric is the most applicable to our analysis. Indeed, 
as we do not include a component for 'sentence importance', 
we can't evaluate relevance properly. 



\subsection{Correlation with Human Judgment} \label{corrwithjugement}

Our initial experiments (Table~\ref{tab:combined_spearman}) help to better understand the strengths and weaknesses of our architecture by comparing the variants presented in Section~\ref{sec:var}.

\noindent 
\textit{Local Context Preservation:} The \textbf{3-Chunk} variant preserves local context by processing three-sentence chunks, improving the consistency correlation from $0.30$ (Base) to $0.39$.\\
Processing text in smaller, semantically coherent chunks can improve the extraction accuracy of statements because dependencies and references are usually resolved within a local context. When processing long documents as a whole, the LLM may struggle to maintain consistency across distant sections, potentially missing key contextual cues that are easier to capture in a smaller scope.

\noindent
\textit{Direct Source Comparison:} The \textbf{StSum\_Text} version implements direct source comparison, further improving consistency correlation to 0.58.\\
Avoiding intermediate representations when matching summary content to source text should reduce information loss because each transformation step introduces potential distortions. Direct comparison preserves the original semantic relationships and contextual nuances present in the source text.

\begin{table}[htb]
\centering
\caption{Spearman Correlation (F1) on SummEval on different pipelines}
\begin{tabular}{lccc}
\textbf{Metrics} & \textbf{ \ Base \ } & \textbf{3-Chunk} & \textbf{StSum\_text} \\ \hline
 Fluency & 0.126 & 0.207 & \textbf{0.351} \\ \hline
 Consistency & 0.231 & 0.306 & \textbf{0.580}\\ \hline
 Coherence & 0.165 & 0.152 & \textbf{0.264} \\ \hline
 Relevance & 0.210 & 0.209 & \textbf{0.300} 
\end{tabular}
\label{tab:combined_spearman}
\end{table}

We explain the substantial improvement from Base to \textit{StSum\_Text} (+0.28 correlation) with the hypothesis that both local context and direct source comparison are important. However, this specialized design shows expected trade-offs: while achieving state-of-the-art consistency correlation, performance on other dimensions (relevance: 0.30, coherence: 0.26, fluency: 0.35) remains moderate. Since the summaries in the SummEval dataset are short, the difference in the number of statements extracted between processing the full text vs. 3-Chunk is negligible. Therefore, we will use the \textbf{StSum\_Text} version for our subsequent analyses.

\subsubsection*{Comparison with Existing Approaches}

Table \ref{tab:benchmark_comparison} reveals several key insights about summarization evaluation approaches:
\\ \noindent
\textit{Traditional Metrics:} N-gram based approaches (ROUGE family) show consistently poor correlation across all dimensions (0.11-0.19), but can be interpretable.
%
\\ \noindent
\textit{Semantic Metrics:} Despite sophisticated embedding spaces, BERTScore and MOVERScore, which are black-box, achieve limited consistency correlation (0.11-0.16).  QuestEval have a significant improvement (0.306) but, based on generated questions and generated answers, the explainability is controversial. 
\\ \noindent
\textit{LLM-based Methods:} While general-purpose LLM evaluators like G-Eval show strong performance across all dimensions (0.52-0.58) but are expensive to run and still black box. Our specialized approach achieves superior consistency correlation (0.58 vs 0.52) with interpretability. 

\begin{table}[h!]
\centering
\renewcommand{\arraystretch}{1.2} 
\caption{Spearman correlation comparison of our approach against other summarization metrics on SummEval dataset}
\small 
\resizebox{\columnwidth}{!}{%

\begin{tabular}{clccccc}
\hline
\textbf{Architecture} & \textbf{Metric} & \textbf{Fluency} & \textbf{Consistency} & \textbf{Coherence} & \textbf{Relevance} & \textbf{Average} \\ \hline
GPT4                             & \textbf{G-Eval (Best)} & \textbf{0.455} & 0.507          & \textbf{0.582} & \textbf{0.547} & 0.523 \\ 
GPT3                             & \textbf{GPTScore}      & 0.403          & 0.449          & 0.434          & 0.381          & 0.417 \\ \hline
\multirow{3}{*}{n-gram}          & \textbf{ROUGE-1}       & 0.115          & 0.160          & 0.167          & 0.326          & 0.192 \\
                                 & \textbf{ROUGE-2}       & 0.159          & 0.187          & 0.184          & 0.290          & 0.205 \\
                                 & \textbf{ROUGE-L}       & 0.105          & 0.115          & 0.128          & 0.311          & 0.165 \\ \hline
\multirow{3}{*}{Embedding based} & \textbf{BERTScore}     & 0.193          & 0.110          & 0.284          & 0.312          & 0.225 \\
                                 & \textbf{MOVERScore}    & 0.129          & 0.157          & 0.159          & 0.318          & 0.191 \\
                                 & \textbf{BARTScore}     & 0.356          & 0.382          & 0.448          & 0.356          & 0.385 \\ \hline
\multirow{2}{*}{T5}              & \textbf{QuestEval}     & 0.228          & 0.306          & 0.182          & 0.268          & 0,246 \\
                                 & \textbf{UniEval}       & 0.449          & 0.446          & 0.575          & 0.426          & 0.474 \\ \hline
qwen2.5:72b                      & \textbf{SEval-Ex}  & 0.351          & \textbf{0.580} & 0.264          & 0.300          & 0.373 \\ \hline
\end{tabular}
}
\label{tab:benchmark_comparison}
\end{table}

These results validate our key design principle: by focusing specifically on consistency evaluation and implementing direct source comparison, we can surpass even sophisticated LLM-based approaches on this crucial dimension. The lower performance on other aspects is an expected trade-off of this specialized design, suggesting that comprehensive summarization evaluation may require combining multiple specialized metrics.

\subsection{Hallucination Detection Analysis}

An hallucination is an AI generative content that appears plausible but is not supported by facts\cite{huang2023survey}, it can happen with all LLM. As hallucinations represent a critical quality issue in generated text, a reliable consistency evaluation metric should show sensitivity to their presence, demonstrating lower correlation scores when hallucinations are present in the summary. 
To evaluate our metric's robustness against different types of hallucinations, we conducted a systematic analysis using the SummEval dataset. We developed three distinct categories of synthetic hallucinations to test our framework's ability to detect these inconsistencies. 

\subsubsection{Hallucination Types}
\begin{figure}[t]
    \centering
    \includegraphics[width=\textwidth]{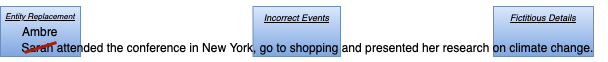}
    \caption{Examples of hallucinations divide in 3 types: Entity Replacement, Incorrect Events and Fictitious Details.}
    \label{fig:exemple_hallucinations}
\end{figure}

Our goal is to establish that summaries that containin hallucinations receive lower scoreaccording toording to our metric. To this end, we designed and implemented three distinct types of hallucinations, illustrated in Figure \ref{fig:exemple_hallucinations}. 

\begin{enumerate}
    \item \textbf{Entity Replacement}: Systematic substitution of named entities with incorrect ones while maintaining the overall structure of the summary.
    
    \item \textbf{Incorrect Events}: Modification of the sequence of events by introducing false temporal or causal relationships. This type of hallucination preserves the entities, but distorts the narrative flow and factual sequence of events. 
    
    \item \textbf{Fictitious Details}: Addition of plausible but unsupported details to the existing summary. This represents a more subtle form of hallucination in which the core information remains intact but is embellished with unsupported details.
\end{enumerate}

\begin{figure}[b!]
    \centering
    \includegraphics[width=0.7\linewidth]{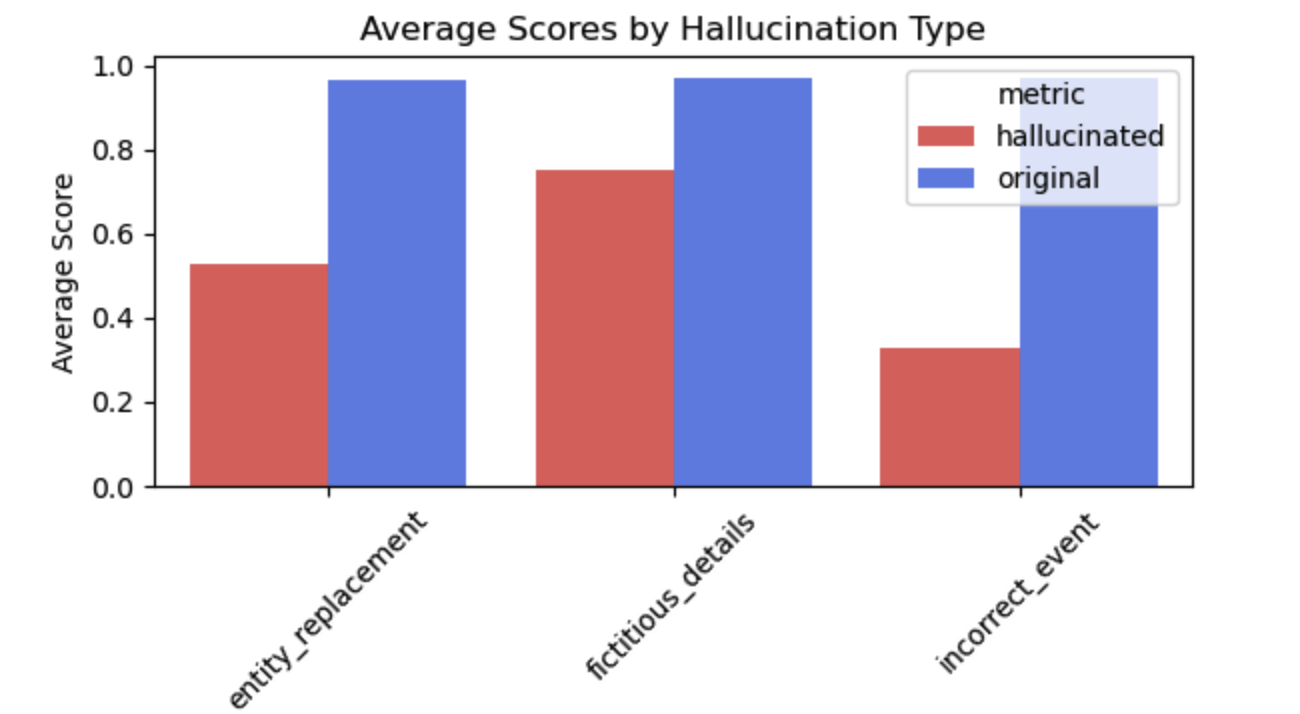}

    \caption{Comparison of average metric scores across different hallucination types, showing the impact on SEval-Ex score.}
    \label{fig:hallucination_comparison}
\end{figure}
\subsubsection{Dataset Preparation}
We used the SummEval dataset as our foundation, creating a balanced dataset of 1,600 samples. The dataset was evenly split into three equals groups per hallucination type. For each summary, we generated a hallucinated version, based on one hallucination type, using controlled prompting through an LLM. The hallucinated dataset is available on the Github.

\subsubsection{Results Analysis}
Our analysis revealed distinct patterns in how our metric responds to different types of hallucinations (Figure \ref{fig:hallucination_comparison}):

\noindent
\textit{Entity Replacement}: Showed a moderate impact with a mean score reduction of 0,435 in correctness score (from 0.964 to 0.529). This drop can be explained because, within an atomic statement, a different event changes its entire meaning and has repercussions on the verdict reasoning.  
\\ \noindent
\textit{Incorrect Events}: Demonstrated the most severe impact, with a 0.65 score reduction in correctness (from 0.970 to 0.328). This is the same explanation as Entity Replacement.
%
\\ \noindent \textit{Fictitious Details}: Exhibited a smaller but still significant impact, with a 0,223 score decrease in correctness (from 0.973 to 0.750). As we only add a little bit noise, the score drop only a little bit because the rest of the summary still good.

All differences were statistically significant (p < 0.0001), confirming our metric's robust ability to detect various forms of hallucination. The varying magnitudes of score reduction across different hallucination types suggest that our metric is particularly sensitive to structural modifications of event sequences while maintaining appropriate tolerance for minor elaborations.



\section{Discussions and Limitations}
Our experimental results demonstrate that StEval-Ex successfully bridges the gap between performance and interpretability in summarization evaluation, particularly for consistency assessment. However, several important limitations and considerations warrant discussion.

\noindent\textbf{Computational Efficiency:} While our use of lightweight, quantized LLMs makes the framework more accessible, the two-stage process (statement extraction followed by verification) increases computational overhead compared to simpler metrics like ROUGE. This trade-off between computational cost and evaluation quality needs to be considered in practical applications.

\noindent\textbf{Prompt Sensitivity:} The framework's performance depends significantly on the quality of statement extraction and verification prompts. While we've optimized these through extensive testing, the sensitivity to prompt design suggests that further improvements might be possible through more sophisticated prompt engineering techniques. To ensure reproducibility, all prompts will be published on the GitHub repository associated with the article.


\noindent\textbf{Scope of Evaluation:} While SEval-Ex excels at consistency evaluation, its performance on other dimensions, particularly on relevance, suggests that based on our experiments, we need to incorporate new types of information to evaluate whether selected sentences are the most relevant sentences from the original text.

\section{Conclusion}
Our work presents SEval-Ex, a novel framework for evaluating summarization quality with interpretable capabilities while providing state-of-the-art correlation with human judgment on consistency assessment. Our key contributions are:

\begin{itemize}
   \item A novel statement-based evaluation framework that decomposes summarization evaluation into fine-grained statement analysis, providing both high correlation with human judgment and interpretable feedback. Our method achieves state-of-the-art correlation (0.580) on consistency assessment.
   
   \item Development of a comprehensive hallucination detection methodology
       
   \item A practical implementation using lightweight, quantized LLMs, making our approach accessible for research and practical applications while maintaining robust performance across varying text lengths.
   
\end{itemize}

These achievements suggest that the decomposition of the summary evaluation into atomic statements offers a promising direction for developing more transparent and effective evaluation methods. As summarization systems continue to advance, such interpretable evaluation frameworks will become increasingly crucial to ensuring their reliability and facilitating their improvement.

Future work could explore extending this approach to better capture document-level properties, such as sentence importance, and investigating ways to optimize computational efficiency while maintaining performance.

%
%
%
\bibliographystyle{splncs04}
\bibliography{references}

\end{document}